\PassOptionsToPackage{unicode}{hyperref}
\PassOptionsToPackage{hyphens}{url}
\PassOptionsToPackage{dvipsnames,svgnames,x11names}{xcolor}
\documentclass[
  10pt,
]{article}
\title{DKnown-Jev-2: Article-Level Answer-Evidence Classification and
Efficient Inference with a Pre-trained Language Model}
\author{Qi Li\thanks{Contact: \href{mailto:liqi@czkj1010.com}{\texttt{liqi@czkj1010.com}}.} \quad Jianjun Xu \quad Jiu Li \quad Pingtao Wei\\Beijing Caizhi Tech, Beijing, China}
\date{}

\usepackage{amsmath,amssymb}
\usepackage{lmodern}
\usepackage{iftex}
\ifPDFTeX
  \usepackage[T1]{fontenc}
  \usepackage[utf8]{inputenc}
  \usepackage{textcomp} % provide euro and other symbols
\else % if luatex or xetex
  \usepackage{unicode-math}
  \defaultfontfeatures{Scale=MatchLowercase}
  \defaultfontfeatures[\rmfamily]{Ligatures=TeX,Scale=1}
\fi
\IfFileExists{upquote.sty}{\usepackage{upquote}}{}
\IfFileExists{microtype.sty}{% use microtype if available
  \usepackage[]{microtype}
  \UseMicrotypeSet[protrusion]{basicmath} % disable protrusion for tt fonts
}{}
\makeatletter
\@ifundefined{KOMAClassName}{% if non-KOMA class
  \IfFileExists{parskip.sty}{%
    \usepackage{parskip}
  }{% else
    \setlength{\parindent}{0pt}
    \setlength{\parskip}{6pt plus 2pt minus 1pt}}
}{% if KOMA class
  \KOMAoptions{parskip=half}}
\makeatother
\usepackage{xcolor}
\IfFileExists{xurl.sty}{\usepackage{xurl}}{} % add URL line breaks if available
\IfFileExists{bookmark.sty}{\usepackage{bookmark}}{\usepackage{hyperref}}
\hypersetup{
  pdftitle={DKnown-Jev-2: Article-Level Answer-Evidence Classification and Efficient Inference with a Pre-trained Language Model},
  colorlinks=true,
  linkcolor={blue},
  filecolor={Maroon},
  citecolor={Blue},
  urlcolor={blue},
  pdfauthor={Qi Li, Jianjun Xu, Jiu Li, Pingtao Wei},
  pdfcreator={LaTeX via pandoc}}
\usepackage[margin=1in]{geometry}
\usepackage{longtable,booktabs,array}
\AtBeginEnvironment{longtable}{\footnotesize\setlength{\tabcolsep}{3pt}}

\usepackage{calc} % for calculating minipage widths
\usepackage{etoolbox}
\makeatletter
\patchcmd\longtable{\par}{\if@noskipsec\mbox{}\fi\par}{}{}
\makeatother
\IfFileExists{footnotehyper.sty}{\usepackage{footnotehyper}}{\usepackage{footnote}}
\makesavenoteenv{longtable}
\ifLuaTeX
  \usepackage{selnolig}  % disable illegal ligatures
\fi

\begin{document}
\sloppy
\maketitle

\hypertarget{abstract}{%
\section*{Abstract}\label{abstract}}

Evidence selection in normative knowledge question answering requires
determining whether a candidate article contains content that directly
supports an answer under the applicable conditions of the question. We
present DKnown-Jev-2, an article-level answer evidence classifier
designed for evidence integration across multiple paragraphs. The model
jointly encodes question conditions, article metadata, and paragraph
content, and uses a pointer-style classification head to predict Answer
Evidence (AE) and Non-answer Evidence (NE) in a single forward pass. On
a manually annotated test set of 972 evaluation instances, DKnown-Jev-2
achieves 91.77\% accuracy and 91.28\% macro-F1, outperforming Jev by
2.16 and 2.37 percentage points, respectively, while reducing both
evidence misses and non-evidence false positives. The test set contains
317 multi-paragraph instances, with a maximum body length of 9,322
characters. Shared article encoding and non-autoregressive
classification support low-latency inference. These results demonstrate
the effectiveness of the proposed method for article-level evidence
selection in normative knowledge question answering.

\textbf{Keywords:} pre-trained language model; article-level evidence
classification; domain adaptation; pointer network; long-context
processing; inference efficiency

\hypertarget{introduction}{%
\section{Introduction}\label{introduction}}

In normative knowledge applications, determining whether a candidate
text provides an answer or part of an answer is different from measuring
semantic relevance to the question. Normative knowledge is issued,
maintained, or published by a responsible authority, and its
applicability depends on the subject, time, version, and other scope
conditions. A candidate text contributes evidence only when it is valid
under those conditions and provides an explicit basis for the requested
matter. Evidence selection therefore requires joint reasoning over
question conditions, article attributes, and textual content.

Candidate articles in practical systems commonly contain metadata and
multiple body paragraphs. Different paragraphs may jointly provide
eligibility, required materials, and procedural information, while also
containing tables of contents, duplicated text, or background
explanations. Compared with the single-paragraph data used in prior
work, the article-level data considered here requires extracting
evidence from longer text, integrating multiple paragraphs, and
distinguishing complete answers, partial answers, and background
information.

We formulate article-level answer evidence identification as a binary
classification task over a user question, its applicable conditions, and
a candidate article. AE includes a complete answer or an explicit answer
component; NE includes contextual material, inapplicable content, and
material without substantive relevance. Each prediction analyzes all
paragraphs in the input article.

Prior DKnown-Jev combines question conditions, article metadata, and a
single candidate paragraph under the same AE/NE label semantics {[}1{]}.
This work extends the prediction unit to all paragraphs belonging to one
article and evaluates article-level classification on 972 instances. We
study overall classification performance and performance stratified by
the number of input paragraphs.

\hypertarget{related-work}{%
\section{Related Work}\label{related-work}}

Retrieval-augmented generation uses external material to support
knowledge-intensive tasks {[}2{]}. Dense retrieval and cross-encoder
reranking provide complementary mechanisms for candidate recall and
relevance assessment {[}3,4{]}. Our task uses answer-support relations
and applicability conditions as the classification criteria.

Natural-language label definitions can participate in classification
decisions, while textual-entailment classification and text-to-text
learning provide related modeling perspectives {[}5,6{]}. We describe AE
and NE with natural-language definitions and determine the label from
matching scores between candidate representations and a decision
representation. Parameter-efficient adaptation methods such as LoRA
provide a practical route for domain adaptation {[}7{]}.

\hypertarget{task-definition-and-label-semantics}{%
\section{Task Definition and Label
Semantics}\label{task-definition-and-label-semantics}}

\hypertarget{input-representation-and-classification-objective}{%
\subsection{Input Representation and Classification
Objective}\label{input-representation-and-classification-objective}}

Each instance is represented as \(x=(q,c,a,P)\), where \(q\) is the user
question, \(c\) contains available conditions such as the service region
and consultation date, \(a\) contains article attributes such as title,
region, and publication date, and \(P=(p_1,ldots,p_m)\) is the ordered
sequence of input paragraphs. The model estimates \(p_\theta(y|x)\),
where \(y\in\{\mathrm{AE},\mathrm{NE}\}\). The predicted label is
\(\hat y=\arg\max_y p_\theta(y|x)\).

The same article can receive different labels for different questions.
Classification must jointly inspect paragraph content and article
metadata to verify the applicable region, time, subject, and scope of
the rule.

\hypertarget{label-definitions-and-decision-criteria}{%
\subsection{Label Definitions and Decision
Criteria}\label{label-definitions-and-decision-criteria}}

\begin{longtable}[]{@{}
  >{\raggedright\arraybackslash}p{(\columnwidth - 4\tabcolsep) * \real{0.33}}
  >{\raggedright\arraybackslash}p{(\columnwidth - 4\tabcolsep) * \real{0.33}}
  >{\raggedright\arraybackslash}p{(\columnwidth - 4\tabcolsep) * \real{0.33}}@{}}
\toprule
\begin{minipage}[b]{\linewidth}\raggedright
Label
\end{minipage} & \begin{minipage}[b]{\linewidth}\raggedright
Name
\end{minipage} & \begin{minipage}[b]{\linewidth}\raggedright
Criterion
\end{minipage} \\
\midrule
\endhead
AE & Answer Evidence & The article is applicable under the conditions in
the question and directly answers all or part of it; content with a
broader scope that includes the user's case is also AE. \\
NE & Non-answer Evidence & The article does not provide explicit answer
content applicable to the user, including background, inapplicable
content, or content without substantive relevance. \\
\bottomrule
\end{longtable}

A partial answer may provide an applicable definition, rule, condition,
material requirement, procedure, or concrete fact; the article need not
answer every aspect of the question. Related material may aid background
understanding without supplying answer content. A rule that does not
apply to the user and cannot support any answer component is NE.

For example, consider the question: ``What conditions and materials are
required for free repair of a Model A device?'' A Model B repair policy
cannot serve as evidence for Model A even if the wording is highly
similar. The current Model A policy ``A manufacturing defect occurring
during the warranty period is eligible for free repair'' answers the
condition component and is therefore a partial answer. A valid
supplementary document stating ``the purchase receipt must be attached''
provides another answer component. A paragraph that merely describes
after-sales-service principles is relevant background, not a procedural
basis. These cases illustrate the distinction between inapplicable
content, partial answers, and contextual references.

The decision uses both paragraph content and article metadata to
distinguish complete answers, partial answers, contextual references,
and inapplicable content, then maps them to AE or NE. Multiple
paragraphs may jointly support an answer. The labels encode
applicability and answer contribution under the given conditions; they
are not thresholded semantic-similarity scores.

\hypertarget{model-output}{%
\subsection{Model Output}\label{model-output}}

The model outputs a probability distribution over AE and NE together
with a predicted label. Metrics are computed against the manually
assigned label. In downstream answer generation, completeness, evidence
conflicts, and citation support must still be checked. When scope
information is insufficient to establish applicability, an application
may retain an unresolved status and verify it after additional scope
information is obtained; this status is not an additional label in the
binary evaluation.

\hypertarget{dataset-and-evaluation-protocol}{%
\section{Dataset and Evaluation
Protocol}\label{dataset-and-evaluation-protocol}}

\hypertarget{dataset-size-and-class-distribution}{%
\subsection{Dataset Availability and
Statistics}\label{dataset-size-and-class-distribution}}

The test set contains 972 manually annotated evaluation instances. Each
instance consists of a user question, applicable conditions, a candidate
article, and its label. Article body length is computed as the sum of
the character counts of all input paragraphs.

The evaluation dataset is publicly available on Hugging Face at
\url{https://huggingface.co/datasets/CaiZhiTech/DKnown-Jev-2}.
The repository provides the 972 evaluation instances in
\texttt{test.jsonl}.

\begin{longtable}[]{@{}lr@{}}
\toprule
Statistic & Value \\
\midrule
\endhead
Evaluation instances & 972 \\
Answer Evidence (AE) & 362 (37.24\%) \\
Non-answer Evidence (NE) & 610 (62.76\%) \\
Single paragraph & 655 (67.39\%) \\
2--3 paragraphs & 191 (19.65\%) \\
4 or more paragraphs & 126 (12.96\%) \\
Median body length & 400.5 characters \\
Maximum body length & 9,322 characters \\
\bottomrule
\end{longtable}

Because NE accounts for 62.76\% of the test set, an always-NE majority
baseline has 62.76\% accuracy. Macro-F1 and per-class metrics are also
reported to evaluate both labels.

\hypertarget{input-alignment-and-evaluation-integrity}{%
\subsection{Input Alignment and Evaluation
Integrity}\label{input-alignment-and-evaluation-integrity}}

We compare DKnown-Jev-2 with Jev and OpenAI Decision. OpenAI Decision
denotes the decision service accessed through the OpenAI Decisions API
{[}8{]}, which selects from predefined candidates and returns a
probability distribution. All three systems predict the same 972 test
instances, yielding 2,916 predictions. Predictions are aligned by
question, applicable conditions, and article content; each instance has
one reference label and three model predictions.

\hypertarget{task-complexity-and-multi-paragraph-long-context-inputs}{%
\subsection{Task Complexity and Multi-paragraph Long-context
Inputs}\label{task-complexity-and-multi-paragraph-long-context-inputs}}

The dataset is substantially more demanding than the prior
single-paragraph setting. Its difficulty arises from evidence
identification under applicability constraints: users may ask about
several matters at once, answer clues may be distributed across
paragraphs, and valid evidence may be interleaved with tables of
contents, duplicated text, and incomplete sentences. The model must
locate and integrate answer components in longer text and then verify
applicability using regional, temporal, and article metadata.

Multi-paragraph inputs account for 32.61\% of the test set, including
12.96\% with at least four paragraphs. These inputs require joint
reasoning over multiple paragraphs to determine whether the article
provides applicable answer evidence.

\hypertarget{model-architecture-and-inference-mechanism}{%
\section{Model Architecture and Inference
Mechanism}\label{model-architecture-and-inference-mechanism}}

\hypertarget{backbone-and-input-organization}{%
\subsection{Backbone and Input
Organization}\label{backbone-and-input-organization}}

DKnown-Jev-2 is built on Qwen3.5-9B {[}9{]} and uses a pointer-style
decision structure. Hidden states at candidate markers and a decision
marker are used by the pointer head to compute AE/NE scores. The input
contains the question and conditions, article metadata, paragraph
sequence, a classification instruction, and natural-language definitions
of AE and NE. A candidate marker terminates each candidate definition,
and a decision marker terminates the input sequence. Paragraphs are
serialized in order with their title and body; article attributes are
shared context. Marker indices are recorded during tokenization to
locate the final hidden states.

\hypertarget{pointer-style-classification-head}{%
\subsection{Pointer-style Classification
Head}\label{pointer-style-classification-head}}

Pointer Networks associate predictions with input positions through
attention scores {[}10{]}. We adopt this position-selection idea by
extracting a representation at each candidate marker and matching it
with the decision representation. Let \(h_k\) be the final hidden state
at candidate marker \(k\) and \(h_{dec}\) the state at the decision
marker:

\[c_k=W_c\mathrm{LN}(h_k),\qquad r=W_q\mathrm{LN}(h_{dec}).\]

With projection dimension \(d=256\), the score is

\[z_k=\frac{c_k^\top r}{\sqrt d}+w^\top c_k.\]

Here \(W_c,W_q\in\mathbb{R}^{d\times H}\) are trainable projections,
\(H\) is the backbone hidden size, and \(w\in\mathbb{R}^d\) is
trainable. Scores are normalized as

\[p_k=\frac{\exp(z_k/\tau)}{\sum_j\exp(z_j/\tau)},\quad k,j\in\{\mathrm{AE},\mathrm{NE}\},\]

where \(\tau>0\) is a temperature parameter. The prediction is
\(\hat y=\arg\max_k p_k\). Both labels share the backbone and
classification head; their semantics are supplied through the
natural-language definitions.

\hypertarget{hybrid-attention-and-cross-paragraph-interaction}{%
\subsection{Hybrid Attention and Cross-paragraph
Interaction}\label{hybrid-attention-and-cross-paragraph-interaction}}

In full-attention layers, the region from the beginning of the decision
context to the last decision marker is bidirectionally visible, allowing
candidate representations to use article context and compare candidates.
Linear-attention layers remain causal. The two attention mechanisms
jointly process question conditions, article paragraphs, and candidate
definitions. Consistent visibility rules are used during training and
inference, allowing candidate and decision representations to integrate
evidence distributed across paragraphs.

\hypertarget{training-objective-and-domain-adaptation}{%
\subsection{Training Objective and Domain
Adaptation}\label{training-objective-and-domain-adaptation}}

Domain adaptation uses supervised training instances labeled AE or NE.
The pointer head outputs a binary probability distribution, and the
one-hot reference label is represented by target vector \(t\). The basic
objective is cross-entropy:

\[L_{CE}=-\sum_k t_k\log p_k.\]

The training framework also supports an RLCD auxiliary objective that
perturbs class scores and optimizes probability-distribution scores.
When enabled, the total objective is \(L=L_{CE}+\lambda L_{RLCD}\),
where \(\lambda\) is the auxiliary-loss weight. We do not attribute
performance gains to RLCD alone.

\hypertarget{non-autoregressive-classification-inference}{%
\subsection{Non-autoregressive Classification
Inference}\label{non-autoregressive-classification-inference}}

Inference constructs the input sequence, records marker positions,
performs one backbone forward pass, extracts candidate and decision
representations, and normalizes classification scores. AE and NE scores
are produced simultaneously and converted with Softmax into a
probability distribution; the maximum-probability label is returned
without autoregressive decoding. Compared with the predecessor's
separate article encoding for each candidate, organizing the article and
all candidate definitions in one sequence reduces repeated encoding and
token-by-token decoding overhead.

\hypertarget{experimental-setup}{%
\section{Experimental Setup}\label{experimental-setup}}

\hypertarget{baselines-and-evaluation-principles}{%
\subsection{Baselines and Evaluation
Principles}\label{baselines-and-evaluation-principles}}

We evaluate DKnown-Jev-2 against Jev and OpenAI Decision on the same
test set. All systems use the same AE/NE semantics and the same manually
annotated reference labels.

\subsection{API Availability}
\label{evaluation-api-access}

An article-relevance API is available for accessing the DKnown-Jev-2
capability. Documentation and request examples are
available at
\url{https://platform.dknowc.cn/static/api-docs/article-relevance.html}.

\hypertarget{metrics}{%
\subsection{Metrics}\label{metrics}}

We report Accuracy, Macro-F1, class precision, and class recall. AE is
the positive class; an NE instance predicted as AE is a false positive
(FP), while an AE instance predicted as NE is a false negative (FN).
\(TP\) and \(TN\) denote correctly identified AE and NE instances, and
\(N\) is the total number of instances.

\[Accuracy=\frac{TP+TN}{N},\quad Precision=\frac{TP}{TP+FP},\quad Recall=\frac{TP}{TP+FN}.\]

\[F1=\frac{2TP}{2TP+FP+FN}.\]

Macro-F1 is the arithmetic mean of the AE and NE F1 scores. Error counts
and class metrics are used to analyze evidence misses and non-evidence
false positives.

\hypertarget{results}{%
\section{Results}\label{results}}

\hypertarget{classification-performance}{%
\subsection{Classification
Performance}\label{classification-performance}}

\begin{longtable}[]{@{}lrrrrrrr@{}}
\toprule
Model & Accuracy & Macro-F1 & AE Precision & AE Recall & NE Recall & FP
& FN \\
\midrule
\endhead
DKnown-Jev-2 & 91.77\% & 91.28\% & 87.11\% & 91.44\% & 91.97\% & 49 &
31 \\
Jev & 89.61\% & 88.92\% & 85.56\% & 86.74\% & 91.31\% & 53 & 48 \\
OpenAI Decision & 83.74\% & 83.00\% & 75.12\% & 84.25\% & 83.44\% & 101
& 57 \\
\bottomrule
\end{longtable}

DKnown-Jev-2 outperforms both baselines on all reported classification
metrics. Compared with Jev, it improves accuracy and Macro-F1 by 2.16
and 2.37 percentage points, respectively, adds 21 correctly classified
instances, and reduces false negatives by 17 and false positives by 4.
This indicates higher AE recall while preserving AE precision.

\hypertarget{complex-article-scenarios}{%
\subsection{Complex Article
Scenarios}\label{complex-article-scenarios}}

We stratify the test set into single-paragraph, 2--3 paragraph, and
4-or-more-paragraph groups.

\begin{longtable}[]{@{}lrrr@{}}
\toprule
Model & 1 paragraph (655) & 2--3 paragraphs (191) & 4+ paragraphs
(126) \\
\midrule
\endhead
DKnown-Jev-2 & 91.45\% & 92.67\% & 92.06\% \\
Jev & 89.16\% & 90.05\% & 91.27\% \\
OpenAI Decision & 83.36\% & 83.25\% & 86.51\% \\
\bottomrule
\end{longtable}

DKnown-Jev-2 has the highest accuracy in all three groups, reaching
92.67\% and 92.06\% in the 2--3 and 4-or-more paragraph groups. The
grouping is based on paragraph count; performance is also affected by
question content, class distribution, and applicability conditions.

In the inference benchmark, on an NVIDIA GeForce RTX 4090 with 24 GB of
VRAM, the model has an average response time of approximately 118 ms with
concurrency set to one, providing low-latency evidence selection.

\hypertarget{conclusion}{%
\section{Conclusion}\label{conclusion}}

We presented DKnown-Jev-2, an article-level answer evidence classifier
that jointly incorporates question conditions, article metadata, and
multiple paragraphs, and performs non-autoregressive classification with
a pointer-style head. The method shifts evidence selection from topical
relevance toward answer support under explicit applicability conditions
and extends the prediction unit from a single paragraph to an article's
paragraph sequence. Experiments show improved evidence recognition in
complex article scenarios. Shared article encoding and
non-autoregressive classification reduce repeated computation and
provide a practical path to low-latency evidence selection. The model
can serve as an evidence-selection component in retrieval-augmented
generation by determining whether a candidate article supports an answer
under the given conditions.

\hypertarget{references}{%
\section*{References}\label{references}}
\small

{[}1{]} Li Q, Xu J, Wei P. DKnown-Jev: Fine-Tuning and Evaluation of a
JEV-Style Model for Tasks Involving Normative Knowledge{[}EB/OL{]}.
arXiv:2610.24358v1, 2026. https://arxiv.org/abs/2610.24358v1. DOI:
10.48550/arXiv.2610.24358.

{[}2{]} Lewis P, Perez E, Piktus A, et al.~Retrieval-Augmented
Generation for Knowledge-Intensive NLP Tasks{[}C{]}//Advances in Neural
Information Processing Systems, 2020, 33: 9459--9474.
https://arxiv.org/abs/2005.11401.

{[}3{]} Karpukhin V, Oguz B, Min S, et al.~Dense Passage Retrieval for
Open-Domain Question Answering{[}C{]}//Proceedings of EMNLP, 2020:
6769--6781. DOI: 10.18653/v1/2020.emnlp-main.550.
https://aclanthology.org/2020.emnlp-main.550/.

{[}4{]} Nogueira R, Cho K. Passage Re-ranking with BERT{[}EB/OL{]}.
arXiv:1901.04085, 2019. https://arxiv.org/abs/1901.04085.

{[}5{]} Yin W, Hay J, Roth D. Benchmarking Zero-shot Text
Classification: Datasets, Evaluation and Entailment Approach{[}EB/OL{]}.
arXiv:1909.00161, 2019. https://arxiv.org/abs/1909.00161.

{[}6{]} Raffel C, Shazeer N, Roberts A, et al.~Exploring the Limits of
Transfer Learning with a Unified Text-to-Text Transformer{[}J{]}.
Journal of Machine Learning Research, 2020, 21(140): 1--67.
https://www.jmlr.org/papers/v21/20-074.html.

{[}7{]} Hu E J, Shen Y, Wallis P, et al.~LoRA: Low-Rank Adaptation of
Large Language Models{[}C{]}//ICLR, 2022.
https://openreview.net/forum?id=nZeVKeeFYf9. Preprint:
https://arxiv.org/abs/2106.09685.

{[}8{]} OpenAI. Decisions{[}EB/OL{]}. OpenAI API Documentation, accessed
2026-11-16. https://developers.openai.com/api/docs/guides/decisions.

{[}9{]} Qwen Team. Qwen3.5-9B{[}EB/OL{]}. Hugging Face Model Card,
accessed 2026-11-16. https://huggingface.co/Qwen/Qwen3.5-9B.

{[}10{]} Vinyals O, Fortunato M, Jaitly N. Pointer
Networks{[}C{]}//Advances in Neural Information Processing Systems,
2015, 28. https://arxiv.org/abs/1506.03134.

\end{document}